\documentclass[10pt,twocolumn,letterpaper]{article}

\usepackage{cvpr}              % To produce the CAMERA-READY version
\usepackage[dvipsnames]{xcolor}

\definecolor{cvprblue}{rgb}{0.21,0.49,0.74}
\usepackage[pagebackref,breaklinks,colorlinks,citecolor=cvprblue]{hyperref}
\usepackage{multirow}

\def\paperID{*****} % *** Enter the Paper ID here
\def\confName{CVPR}
\def\confYear{2024}

\title{MIPI 2024 Challenge on Nighttime Flare Removal: Methods and Results}

\author{
\textbf{Challenge and Workshop Organizers}\\
Yuekun Dai \quad Dafeng Zhang \quad 
Xiaoming Li \quad Zongsheng Yue \quad Chongyi Li \quad Shangchen Zhou \\ Ruicheng Feng \quad Peiqing Yang  \quad  
Zhezhu Jin \quad Guanqun Liu \quad 
Chen Change Loy  \\
\\
\textbf{Challenge Participants}\\
Lize Zhang \quad Shuai Liu \quad Chaoyu Feng \quad Luyang Wang \quad Shuan Chen \quad Guangqi Shao \\ Xiaotao Wang \quad Lei Lei \quad 
Qirui Yang \quad Qihua Cheng  \quad Zhiqiang Xu \quad Yihao Liu \\ Huanjing Yue \quad Jingyu Yang \quad 
Florin-Alexandru Vasluianu \quad Zongwei Wu \quad George Ciubotariu \\ Radu Timofte \quad
Zhao Zhang \quad Suiyi Zhao \quad Bo Wang \quad Zhichao Zuo \quad Yanyan Wei \\ 
Kuppa Sai Sri Teja \quad Jayakar Reddy A \quad Girish Rongali \quad  Kaushik Mitra \quad
Zhihao Ma \\  Yongxu Liu \quad
Wanying Zhang \quad Wei Shang \quad 
Yuhong He \quad Long Peng \\ Zhongxin Yu \quad Shaofei Luo \quad
Jian Wang \quad Yuqi Miao \quad Baiang Li \quad Gang Wei \\ 
Rakshank Verma \quad Ritik Maheshwari \quad Rahul Tekchandani \quad Praful Hambarde \\ Satya Narayan Tazi  \quad Santosh Kumar Vipparthi \quad Subrahmanyam Murala \quad
Haopeng Zhang \\ Yingli Hou \quad Mingde Yao \quad
Levin M S \quad Aniruth Sundararajan  \quad Hari Kumar A
}

\begin{document}
\maketitle
\begin{abstract}
The increasing demand for computational photography and imaging on mobile platforms has led to the widespread development and integration of advanced image sensors with novel algorithms in camera systems.
However, the scarcity of high-quality data for research and the rare opportunity for in-depth exchange of views from industry and academia constrain the development of mobile intelligent photography and imaging (MIPI).
Building on the achievements of the previous MIPI Workshops held at ECCV 2022 and CVPR 2023, we introduce our third MIPI challenge including three tracks focusing on novel image sensors and imaging algorithms.
In this paper, we summarize and review the Nighttime Flare Removal track on MIPI 2024.
In total, 170 participants were successfully registered, and 14 teams submitted results in the final testing phase.
The developed solutions in this challenge achieved state-of-the-art performance on Nighttime Flare Removal.
%
%A detailed description of all models developed in this challenge is provided in this paper.
More details of this challenge and the link to the dataset can be found at \href{https://mipi-challenge.org/MIPI2024/}{https://mipi-challenge.org/MIPI2024}.
\end{abstract}    
\section{Introduction}
\label{sec:intro}
Lens flare, an optical phenomenon, arises when intense light scatters or reflects within a lens system, manifesting as a distinct radial-shaped bright area and light spots in captured photographs.
In mobile platforms such as monitor lenses, smartphone cameras, UAVs, and autonomous driving cameras, daily wear and tear, fingerprints, and dust can function as a grating, exacerbating lens flare and making it particularly noticeable at night. Thus, flare removal algorithms are highly desired.

Flares can be categorized into three main types: scattering flares, reflective flares, and lens orbs. In this competition, we mainly focus on removing the scattering flares, as they are the most prevalent type of nighttime image degradation. 
Early attempts at scattering flare removal were made by Wu \etal~\cite{how_to}, who proposed a dataset with physically-based synthetic flares and flare photos taken in a darkroom. However, these flares have obvious domain gap with real-captured nighttime flares. To address this issue, Dai \etal~\cite{dai2022flare7k,dai2023flare7k++} propose a new dataset Flare7K++ which is specifically designed for nighttime scenes. 
Additionally, various other efforts have been pursued, including flare removal in multi-light scenarios~\cite{zhou2023improving}, in raw image formats~\cite{lan2023tackling}, and smartphone reflective flare~\cite{dai2023reflective}.
However, due to variations in lens structures and the diversity of lens protectors, existing lens flare datasets struggle to cover all types of lens flare comprehensively. 
This occasionally results in `out of distribution' occurrences of lens flare in real-world captures for specific type of lenses. 
In response to the growing demand among smartphone and lens manufacturers, this competition focuses on developing lens-specific lens flare removal methods.
In addition to the Flare7K++ dataset, we provide 600 aligned flare-corrupted/flare-removed image pairs specifically for certain smartphone's rear camera.
Furthermore, in order to mimic the commonly-used high resolutions in the industry, all training set and test set images' resolutions are set to 2K.

We hold this challenge in conjunction with the third MIPI Challenge which will be held on CVPR 2024. Similar to the previous MIPI challenge~\cite{sun2023mipi,sun2023mipi2,dai2023mipi,zhu2023mipi}, we are seeking an efficient and high-performance image restoration algorithm to be used for recovering flare-corrupted images. MIPI 2024 consists of three competition tracks:

\begin{itemize}
    \item \textbf{Few-shot RAW Image Denoising} is geared towards training neural networks for raw image denoising in scenarios where paired data is limited.
    \item \textbf{Demosaic for HybridEVS Camera} is to reconstruct HybridEVS's raw data which contains event pixels and defect pixels into RGB images.
    \item \textbf{Nighttime Flare Removal} is to improve nighttime image
    quality by removing lens flare effects.
\end{itemize}

\section{MIPI 2024 Nighttime Flare Removal}
\label{sec:track}

To facilitate the development of efficient and high-performance flare removal solutions, we provide a high-quality dataset to be used for training and testing and a set of evaluation metrics that can measure the performance of developed solutions.
This challenge aims to advance research on nighttime flare removal.

\subsection{Datasets}
Our competition provides a paired flare-corrupted/flare-free dataset that contains 600 aligned training images in 2K resolution. Participants can train a pixel-to-pixel network with this dataset for flare removal. The validation set and testing set consist of 50 and 50 pairs of images, respectively. The input images from the validation and testing set are provided and the ground truth images are not available to participants.
In addition, participants can also use Flare7k++~\cite{dai2023flare7k++} as an additional training dataset and its released checkpoint. The Flare7k++ provides 5,000 synthetic flare images in 1440$\times$1440, 962 real flare images in 756$\times$1008, and 23,949 background images. Flare images can be added to the flare-free background images to synthesize paired data for training.

\subsection{Evaluation}
In this competition, we mainly focus on the perceptual similarity of the flare-removed image and flare-free ground truth. 
Thus, we choose to use the Learned Perceptual Image Patch Similarity (LPIPS)~\cite{lpips} as our main evaluation metric. 
Besides, Peak Signal-to-Noise Ratio (PSNR) and Structural Similarity Index Measure (SSIM)~\cite{ssim} are also listed as references. 
Participants can view these metrics of their submission to optimize the model's performance.

\subsection{Challenge Phase}
The challenge consisted of the following phases:
\begin{enumerate}
    \item Development: The registered participants get access to the data and baseline code, and are able to train the models and evaluate their running time locally.
    \item Validation: The participants can upload their models to the remote server to check the fidelity scores on the validation dataset, and to compare their results on the validation leaderboard.
    \item Testing: The participants submit their final results, code, models, and factsheets.
\end{enumerate}
\section{Challenge Results}

Among $170$ registered participants, $14$ teams successfully submitted their results, code, and factsheets in the final test phase.
Out of these, 12 teams have contributed their solutions to this report.
Table \ref{tab:result} reports the final test results and rankings of the teams. 
Only two teams train their models with extra data of real-captured nighttime background images.
The methods evaluated in Table \ref{tab:result} are briefly
described in Section \ref{sec:methods} and the team members are listed in Appendix.
Finally, the MiAlgo\_AI team is the first place winner of this challenge, while BigGuy team win the second place and SFNet-FR team is the third place, respectively.

\begin{table*}
\centering
\caption{Results of MIPI 2024 challenge on nighttime flare removal. ‘Runtime’ for per image is tested and averaged across the validation datasets, and the image size is $1440\times1920$. ‘Params’ denotes the total number of learnable parameters.}
\label{tab:result}
\scalebox{0.7}{
\begin{tabular}{l|ccc|ccccc}
\toprule
\multicolumn{1}{c}{}                                                        & \multicolumn{3}{|c|}{Metric}                                                               & \multicolumn{1}{l}{}                             &                               &                            &                              &                            \\
\multicolumn{1}{l|}{\multirow{-2}{*}{Team Name}} & LPIPS$^*$ & PSNR & SSIM & \multicolumn{1}{l}{\multirow{-2}{*}{Params (M)}} & \multirow{-2}{*}{Runtime (s)} & \multirow{-2}{*}{Platform} & \multirow{-2}{*}{Extra data} & \multirow{-2}{*}{Ensemble} \\
\midrule
MiAlgo\_AI                          & $0.1435_{(\textbf{1})}$  & $22.15_{(1)}$           & $0.7075_{(2)}$        & 141.61             & 8.0      & NVIDIA Tesla A100                & Yes          & -                      \\
BigGuy                               & $0.1502_{(\textbf{2})}$  & $21.50_{(7)}$           & $0.6996_{(7)}$        & 26.13             & 30.0      & NVIDIA RTX 3090               & -            & self-ensemble                        \\
SFNet-FR                               & $0.1518_{(\textbf{3})}$  & $21.74_{(3)}$           & $0.7188_{(1)}$        & 383.42             & 0.017      & NVIDIA RTX 3090Ti                   & -            & -     \\
LVGroup\_HFUT                              & $0.1620_{(4)}$  & $21.71_{(5)}$          & $0.7041_{(4)}$       & /              & 0.055         & NVIDIA RTX 4090                & -            & -                          \\
NativeCV                              & $0.1688_{(5)}$  & $21.39_{(8)}$          & $0.6929_{(8)}$       & /              & /         & /                 & -            & -                          \\
CILAB-IITMadras                        & $0.1697_{(6)}$  & $21.70_{(6)}$           & $0.7042_{(3)}$        & 61.40             & 0.70      & NVIDIA RTX 4090                 & -            & model-ensemble                         \\
Xdh-Flare                           & $0.1703_{(7)}$  & $21.99_{(2)}$           & $0.7005_{(5)}$        & 24.47             & 1.78      & NVIDIA RTX 4090                  & Yes            & -                          \\
Fromhit                                      & $0.1713_{(8)}$  & $21.24_{(9)}$           & $0.6850_{(10)}$        &     /              &  1.256         & NVIDIA RTX A6000                   &   -          & -              \\
UformerPlus                               & $0.1732_{(9)}$  & $21.73_{(4)}$           & $0.6997_{(6)}$        & 38.79              & 0.32      & NVIDIA RTX 3090                & -            & -             \\
GoodGame                                 & $0.1813_{(10)}$  & $20.85_{(10)}$           & $0.6881_{(9)}$        & 19.47              & 0.73      & NVIDIA RTX 3090                  & -            & -              \\
IIT-RPR                           & $0.1926_{(11)}$ & $20.66_{(11)}$           & $0.6775_{(11)}$        & 20.47             & 1.72      & NVIDIA RTX 2080Ti                & Yes            & -              \\
LSCM-HK                  & $0.1926_{(12)}$ & $22.66_{(12)}$          & $0.6775_{(12)}$       &    20.47               &   /       & NVIDIA RTX 3090               &       -       &          -                 \\
Hp\_zhangGeek                  & $0.2332_{(13)}$ & $19.74_{(13)}$          & $0.6538_{(13)}$       &    2.11               &   0.13       & NVIDIA RTX 4090               &       -       &         self-ensemble                 \\
Lehaan                  & $0.6749_{(14)}$ & $16.27_{(14)}$          & $0.4655_{(14)}$       &    /               &   /       & NVIDIA RTX 4050             &       -       &          -                 \\
\bottomrule
\end{tabular}
}
\end{table*}

\section{Methods}
\label{sec:methods}

\begin{figure*}
    \centering
    \includegraphics[width=0.9\linewidth]{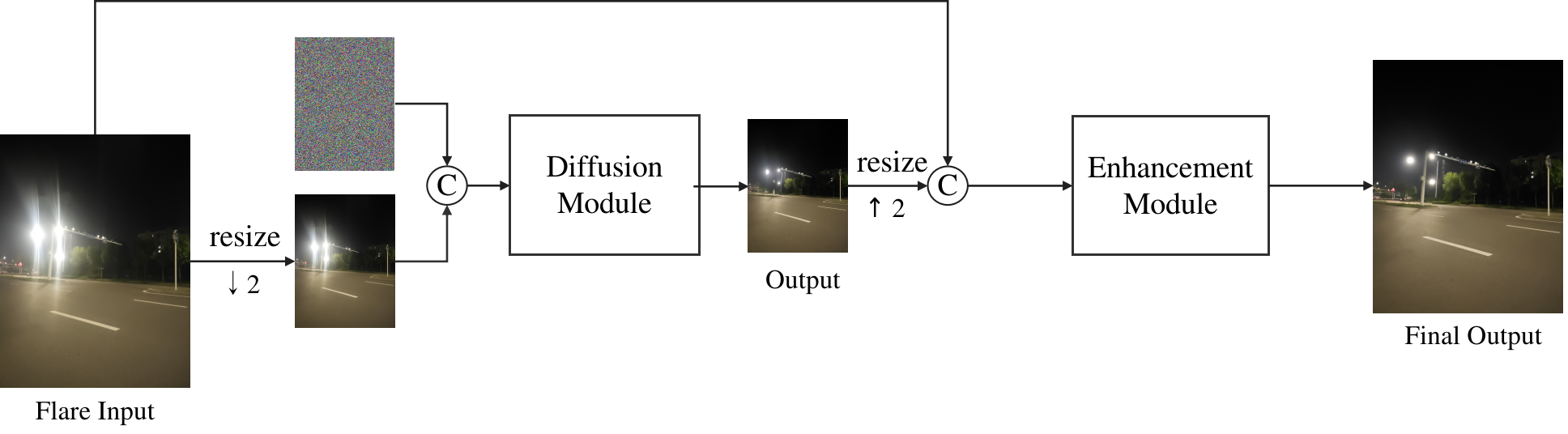}
    \vspace{-3mm}
    \caption{The network architecture of MiAlgo\_AI. }
    \label{fig:PPDN}
    \vspace{-2mm}
\end{figure*}

\paragraph{\bf MiAlgo\_AI} This team proposes a Progressive Perception Diffusion Network (PPDN). By implementing a two-stage network architecture, it generates visually high-quality results in a progressive strategy. Specifically, in the first stage, there is a diffusion module that aims to remove the flares of the input to the greatest extent feasible. Inspired by \cite{luo2023image}, they use the IR-SDE as the base diffusion module, which can avoid generating smooth results during deflaring. In the second stage, they utilize the AOT Block \cite{zeng2022aggregated} as the fundamental enhancement module to amplify the details of the flat domain in the output of the first stage and recover the content of the flare texture. Significantly, the output of the diffusion module serves as crucial conditional information. The output of the inpainting module has high-quality visualization effects. The entire pipeline is shown in Figure \ref{fig:PPDN}.

When generating a training dataset, it contains a synthetic dataset and a real dataset. Borrowing from \cite{dai2023flare7k++}, they additionally collected several high-quality nighttime photos at two resolutions and added compound flares as a synthetic dataset. A paired flare dataset containing 600 aligned training images provided by the event group is the real dataset. Upon examining the validation images, the authors noticed a variation in brightness between the input and output images. Furthermore, to simulate more realistic flare-corrupted images when encountering heavier input fog, they augmented the base image with additional light and local haze

When training, they first train the diffusion module using synthetic data generated online and real dataset for about 400,000 iterations. Then they fix the weight of the diffusion module and train the enhancement module for about 300,000 iterations with a batch size of 4. The initial learning rate is $lr=1e-4$, and cosine annealing is used to reduce the learning rate. To optimize the model's capacity for extracting global information, the team abstains from employing a random cropping strategy on the input image. The training process is executed using the power of 2 Tesla A100 GPUs for approximately 3 days.

\paragraph{\bf BigGuy} This team designs a one-stage Restormer-like Structure \cite{Zamir2021Restormer}, making full use of the hierarchical multi-scale information from low-level features to high-level features. To ease the training procedure and facilitate the information flow, an efficient Transformer for image restoration is utilized to model global connectivity and is still applicable to large images. Since flares can take up a large portion of the image, and possibly the entire image, during the removal of nighttime flares, it is critical to have a large receptive area. However, conventional window-based transformer methods limit the receptive field within the window, thus limiting their ability to capture global features. Also, to allow the network to focus more on the flare region, this team used a difference algorithm to obtain a mask between the input image and the ground truth to compute the loss function.

The training is using the Adam optimizer \cite{kingma2014adam}, with $lr=1e-5$ and default decay parameters. The optimized objective is a mixture of terms, combining the L1 loss, VGG \cite{simonyan2014very} loss, mask loss, and LPIPS loss comparing the output restored image to the reference image.

\begin{figure*}
    \centering
    \includegraphics[width=0.9\linewidth]{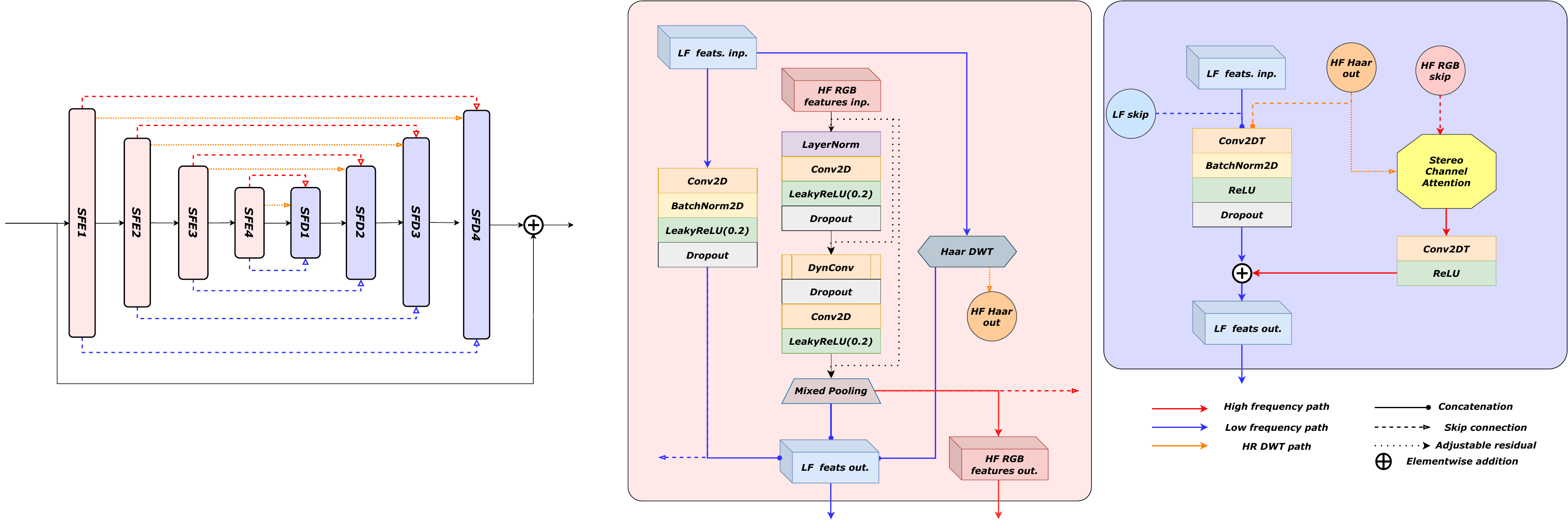}
    \vspace{-3mm}
    \caption{A graphical representation of the proposed SFNet \emph{(left)}, with a detailed representation of the Spatial-Frequency Encoder (SFE) \emph{(center)}, and the Spatial-Frequency Decoder (SFD) \emph{(right)}. }
    \label{fig:SFNet}
    \vspace{-2mm}
\end{figure*}

\paragraph{\bf SFNet-FR}
% Network structure
This team proposes \emph{SFNet}, a solution based on multiple-level frequency-band decomposition, performed in both the RGB spatial domain and the image frequency domain. Figure \ref{fig:SFNet} offers a graphical representation of the \emph{SFNet} model, with the encoder module \emph{(center)} and the paired decoder \emph{(right)}. The solution follows the UNet structure \cite{ronneberger2015u}, with multiple frequency-band skip connections, such as the high-frequency RGB domain \emph{(red)}, or Haar Discrete Wavelet Transform (DWT) \emph{(orange)} skip connection, with another preserving low-frequency combined domains features \emph{(blue)}.   

The information preserved through the aforementioned skip connections is the output of the dual domain splitting performed at the encoder level. In the RGB domain, the splitting is performed through a module combining the \emph{AvgPooling} and the \emph{MaxPooling} operators, while the splitting in the frequency domain is done through a Haar DWT operator.  The solution builds on previous work \cite{9897883}, with a considerable model complexity allocated to the modules processing high-frequency information, while the lower complexity features are refined through simpler modules. 

At the decoder level, the multi-domain high-frequency information is fused in a Stereo Channel Attention module \cite{chen2022simple}, while the low-frequency features are processed separately, then receiving the enhanced high-frequency information as compensation. 

The training is using the Adam optimizer \cite{kingma2014adam}, with $lr=2e-4$ and default decay parameters. The inputs are cropped into $320\times320$ with a batch size of 1. The optimized objective is a mixture of terms, combining the L1 loss with a VGG \cite{simonyan2014very} loss, and a gradient loss comparing the Sobel gradient of the output restored image to the gradient of the reference image. 

This builds SFNet as a capable solution for the flare removal task which, trained on the Flare7K++ dataset and the challenge data, can achieve a significant performance level with consistent results for all the evaluated metrics. As a single-stage solution, solving the nighttime flare removal in an end-to-end fashion (without using any self-ensemble or model-ensemble structures), SFNet represents a good trade-off in terms of achieved performance for the characteristic computational cost. The solution is one of the fastest compared to the other competitors, being able to perform real-time flare removal on a consumer-grade GPU, the Nvidia RTX 3090Ti. 

\begin{figure}
    \centering
    \includegraphics[width=1.0\linewidth]{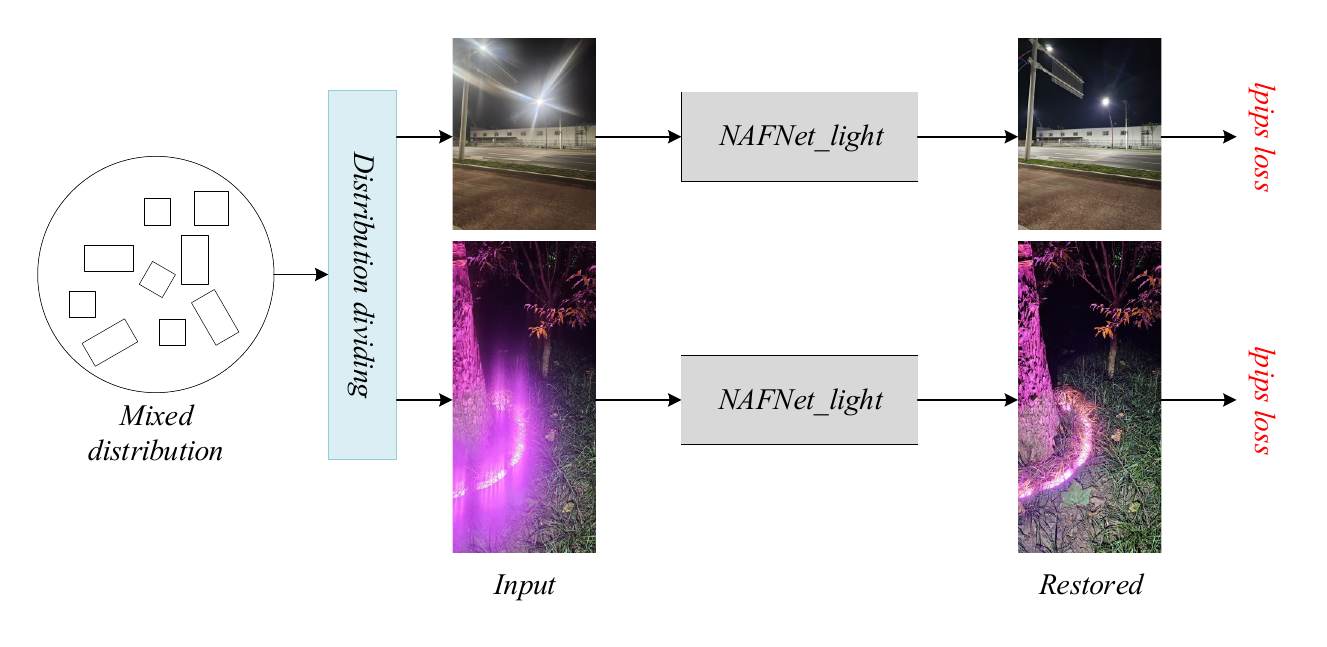}
    \vspace{-6mm}
    \caption{The network architecture of LVGroup\_HFUT team. }
    \label{fig:Arch_LVGroup_HFUT}
    \vspace{-2mm}
\end{figure}

\paragraph{\bf LVGroup\_HFUT}
This team proposes to divide the whole training dataset into different subsets based on different distributions, then train the neural network model on each subset separately, and finally integrate the obtained results to realize flare removal. Specifically, this team refers to NAFNet \cite{chen2022simple}  and FCL-GAN \cite{fclgan} to construct the model, and then divide the provided training data into two subsets (according to the resolution due to the difference in their distributions) and train the models separately to obtain two pre-trained models, Fig.\ref{fig:Arch_LVGroup_HFUT} demonstrates the detailed architecture of this team. 

\textit{Training description.} The proposed architecture of this team is based on PyTorch 2.2.1 and an NVIDIA 4090 with 24G memory. They set 2500 epochs for training with batch size 4, using AdamW with $\beta_1$=0.9 and $\beta_2$=0.999 for optimization. The initial learning rate was set to 0.0002, which was reduced by half every 50 epochs. For data augment, they first randomly crop the image to 768$\times$768 and then perform a horizontal flip with probability 0.5. Besides, as mentioned above, two different models were trained separately to fit different distributions (different resolutions in the experiment).

\textit{Test description.}
Similarly to the training stage, the test image with the original resolution is fed into the two models according to their distributions for inference to obtain the results, and finally the obtained results are combined.

\begin{figure*}
    \centering
    \includegraphics[width=0.9\linewidth]{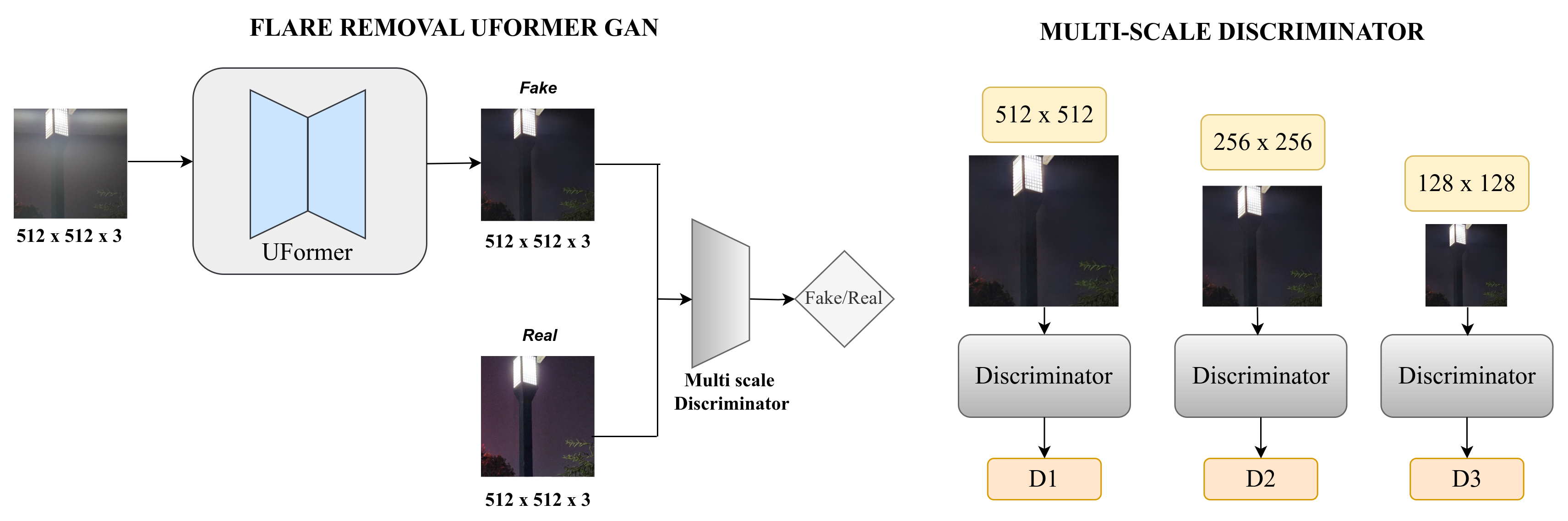}
    \vspace{-3mm}
    \caption{Overview of FRU-GAN Architecture with multi-scale discriminator. }
    \label{fig:frugan}
    \vspace{-2mm}
\end{figure*}

\paragraph{\bf CILAB-IITMadras}
This team proposes to ensemble 3 Uformers using different metrics and methodologies. Flare Removal Uformer GAN(FRUGAN) utilizes UFormer\cite{wang2022uformer} as a generator and a multiscale discriminator that utilizes both adversarial and feature-matching loss to remove flare from the given image. They have used three discriminators similar to the network in pix2pixhd\cite{wang2018pix2pixHD} operating at different image scales as shown in 
Fig.\ref{fig:frugan}. 

\begin{figure}
    \centering
    \includegraphics[width=0.9\linewidth]{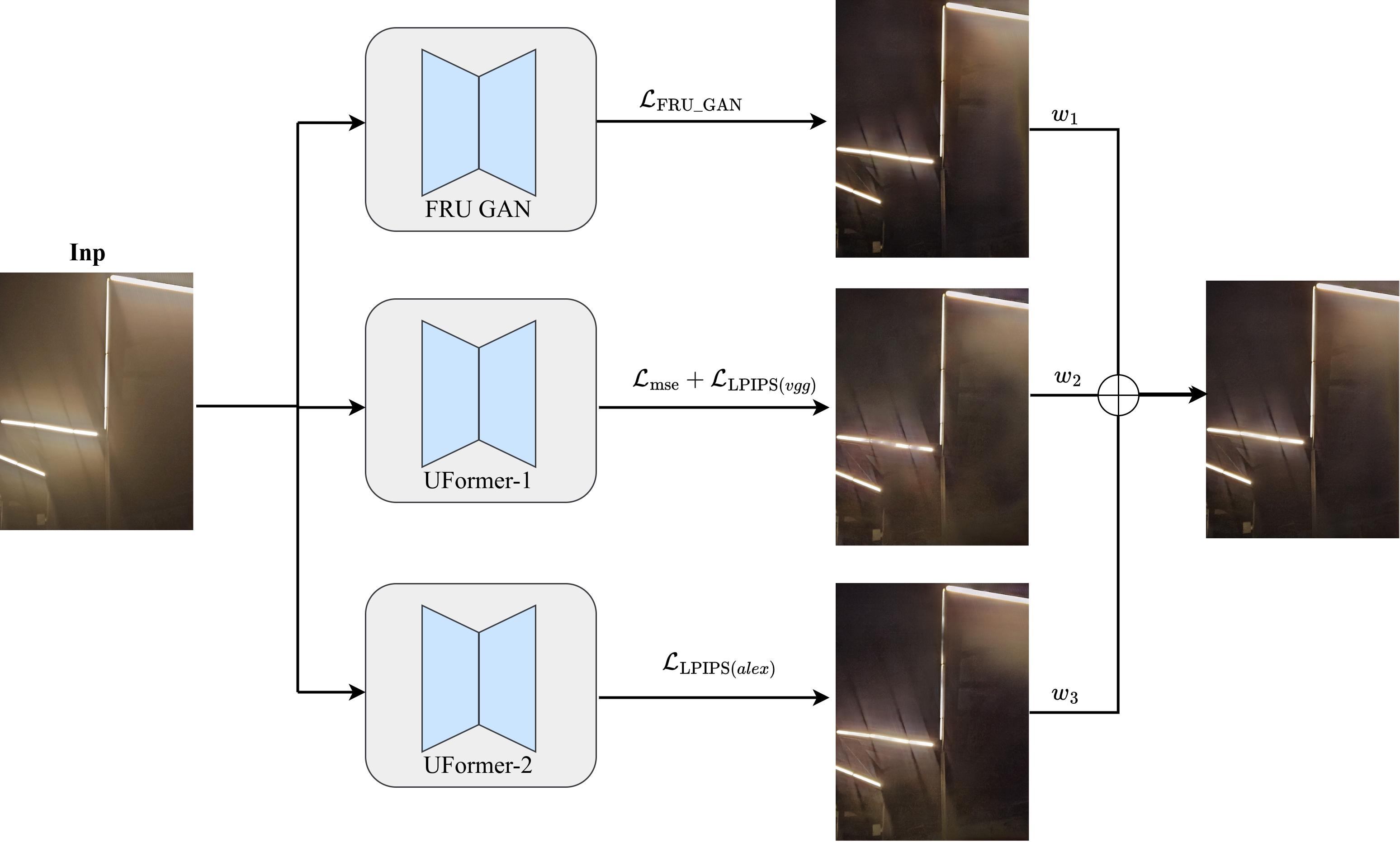}
    \vspace{-1mm}
    \caption{Ensemble Model with blended output. }
    \label{fig:ensemble}
    \vspace{-2mm}
\end{figure}

A combination of adversarial loss, multiscale discriminator loss,  L1, and perceptual losses is used to train FRUGAN. To improve the results of FRUGAN, Uformer-1, and Uformer-2 were trained on losses $L_{mse}+L_{LPIPS}$ and $L_{LPIPS}$ respectively. The complete ensemble model uses weights $w1=0.60$, $w2=0.25$, and $w3=0.15$ as depicted in Fig.\ref{fig:ensemble}. 

They have trained the proposed Uformer model by randomly cropping the images from the competition dataset to 800*800 images and then resizing it to 512$\times$512 images. They further did data augmentations of random horizontal and vertical flips and random rotations of up to 5$^{\circ}$. The model was trained with Adam optimizer with $\beta_{1}$ set to 0.9 and $\beta_{2}$ set to 0.999. This team found that the model started over-fitting after 50 epochs. Both the discriminator and the generator were updated after every iteration. 
The FRUGAN model was trained on NVIDIA A100 with 40GB VRAM. 

Similarly, the other Uformer models were trained by resizing the input images to 512*512. The model was trained with Adam optimizer with $\beta_{1}$ set to 0.9 and $\beta_{2}$ set to 0.999. The learning rate was set to 1e-4. They have trained for 100 epochs. The Uformer models were trained on NVIDIA RTX 4090 with two 24GB VRAM GPUs.

\paragraph{\bf Xdh-Flare}
This team adopts Uformer architecture \cite{Uformer} and makes several improvements in the dataset and loss function to remove nighttime flare. The authors observe that the data distribution in the target domain differs from that in the source domain, primarily manifesting in the discrepancy between the proportion of light sources and flare tones training set images and the testing set. In response to the data disparity between the target and source domains, the authors adopt a strategy of augmenting the dataset in the target domain to reduce domain gaps. Following the method of synthesizing data from Flare7k++ \cite{dai2023flare7k++}, they add flare images provided by Flare7k++ to the flare-free background images provided by BracketFlare \cite{dai2023reflective} to synthesize paired data for training. They observe the flare tones of all images in the dataset around light sources, selecting flare images based on a similar distribution of flare tones in the flare images, completing the augmentation of the dataset from the source domain to the target domain.

They use L1 loss, SSIM loss, and perceptual loss to remove flare regions. Given an image with flares \( I_{input} \), Uformer network outputs flare-free image \( I_g \) and flare image \( I_f \). To better recover the flare-free image and flare image, the authors add \( I_f \) and \( I_g \) , then calculate the L1 loss with \( I_{input} \). The total L1 loss is represented as:
\begin{equation}
    \text{loss}_{L1} = \| I_{input} - (I_g + I_f) \| * w + \| I_o - I_g \|
\end{equation}
where \( w \) is a hyper-parameter set to be 2.

Using the SSIM loss function can make the generated images closer to the real images in terms of structure, improving the quality of the generated images. SSIM Loss is defined as:
\begin{equation}
    \text{loss}_{SSIM} = 1 - \text{SSIM}(I_o, I_g)
\end{equation}

Using perceptual loss can measure perceptual similarity, improving the similarity between the output images and input images. \( f(\cdot) \) represents the perceptual extraction network, and they use AlexNet pretrained on ImageNet to extract features:
\begin{equation}
    \text{loss}_{per} = \| f(I_o) - f(I_g) \| * w + \| f(I_g + I_f) - f(I_{input}) \|
\end{equation}
where \( w \) is a hyper-parameter set to be 2.

The total loss is the weighted sum of the three types of loss:
\begin{equation}
    \text{loss}_{total} = \alpha * \text{loss}_{L1} + \beta * \text{loss}_{SSIM} + \gamma * \text{loss}_{per}
\end{equation}
where \( \alpha, \beta, \) and \( \gamma \) are respectively set to 1, 0.01, and 1 in their experiments.

When training, the input images are cropped to \( 512 \times 512 \). The authors also use gamma correction and inverse gamma correction to the input images. The authors use the Adam optimizer with an initial learning rate of 0.0001 and a batch size of 4. The network is trained for 500 epochs on the augmented dataset.
\paragraph{\bf Fromhit}
%Network structure
This team employs an efficient image restoration model, NAFNet\cite{chen2022simple}, as the base model. 
Specifically, a four-scale CNN encoder and decoder are adopted, and each scale contains two NAFBlocks. Between the encoder and decoder, the authors use four NAFBlocks as a middle block. Then, the authors design a loss function for nighttime flare removal. During training, the authors minimize the sum of two losses, $L_1$ loss encourages the predicted flare-free image to be close to the ground truth both photometrically and perceptually. Like \cite{wu2021train}, the perceptual loss is computed by feeding the predicted flare-free image and ground-truth through a pre-trained VGG-19 network\cite{simonyan2014very}. This team does not process light sources.
%
%Training details
For training, the authors randomly cropped $512\times512$ patches from the training images as inputs. The mini-batch size is set to 4 and the whole network is trained for $1\times10^5$ iterations. 
The learning rate is initialized as $1\times10^-4$, and the authors use ADAM as the optimizer with $\beta_{1} = 0.9$, and $\beta_{2} = 0.99$,

\paragraph{\bf UformerPlus}
The team proposes an effective nighttime flare removal pipeline. Firstly, they employed the strong image restoration model Uformer \cite{Uformer} as the base model, which has an encoder, a decoder, and skip connections. The Locally-enhanced Window (LeWin) block is adopting the design in Uformer. And then to leverage the frequency characteristics of the image, the authors introduce the ResFFTBlock \cite{mao2021deep} after the LeWin block, which is based on Fast Fourier Convolution (FFC), to extract global frequency features for reducing distortions and enhancing details. Moreover,  the authors use two NAFBlocks \cite{chen2022simple} as the refinement module following the last decoder blocks for powerful representation. Finally, some improvements were made to the loss function.  Instead of using a fixed loss weight, the team dynamically adjusts the weight ratio of the loss function as the training iterations progress, with an increased emphasis on perceptual loss for better visual results during training. Through model fusion and weighted loss function, the performance of the model was further improved and ultimately achieved competitive results in the challenge.

\begin{figure*}
    \centering
    \includegraphics[width=0.9\linewidth]{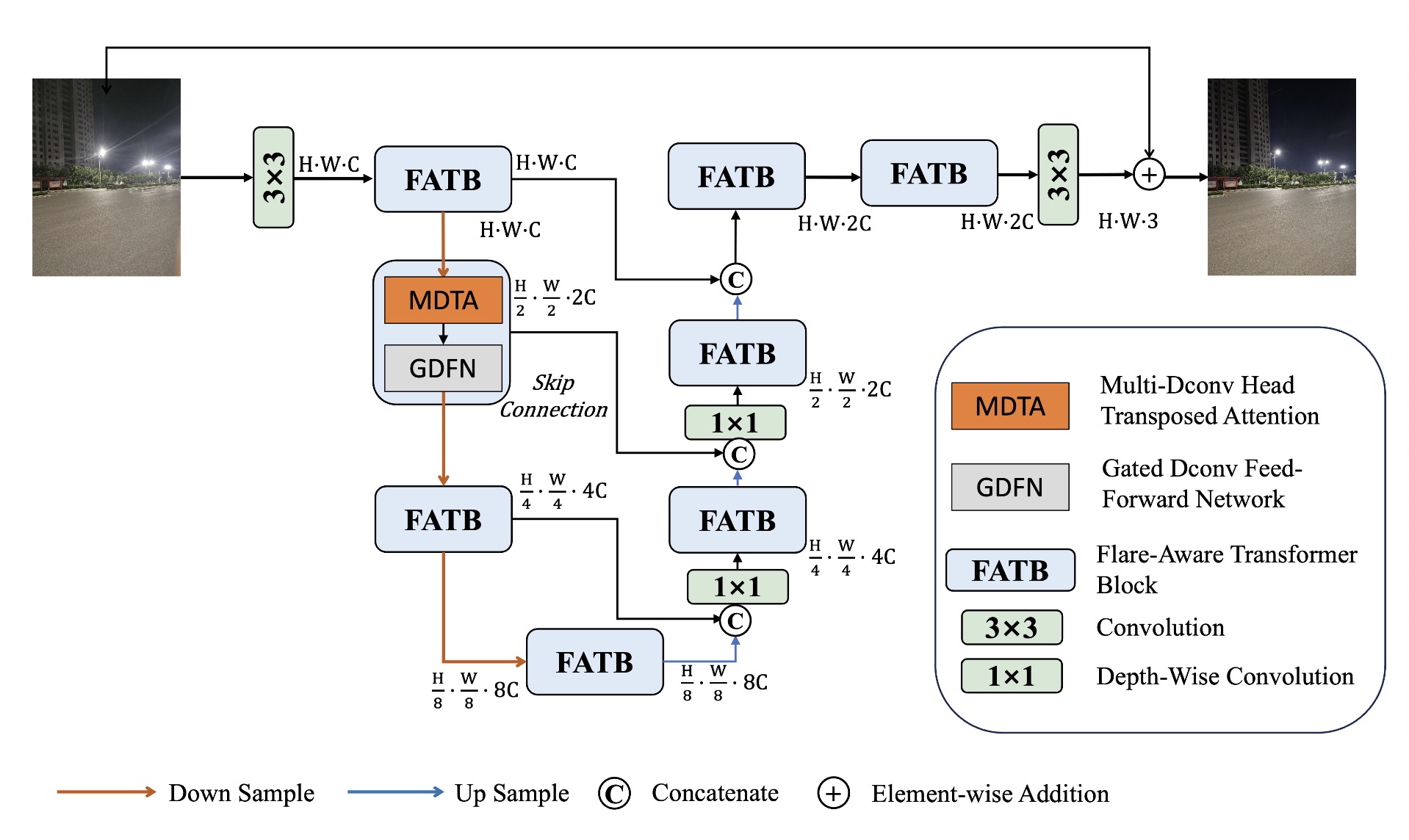}
    \vspace{-3mm}
    \caption{The network architecture of GoodGame team. MDTA and GDFN are the same as Restormer\cite{Zamir2021Restormer}.}
    \label{fig:GoodGame}
    \vspace{-2mm}
\end{figure*}

The loss function comprises both  Charbonnier $L_1$ loss and perceptual loss, with dynamically assigned weights. The inputs are cropped into $512\times512$ with a batch size of 2, and the Adam optimizer \cite{kingma2014adam} is used. The initial learning rate is set to $1 \times 10^{-4}$, and the CosineAnnealingLR scheduler is employed with a maximum of 300,000 iterations and a minimum learning rate of $1 \times 10^{-6}$ to adjust the learning rate.  They also use horizontal and vertical flips for data enhancement. For testing, the authors split the original images into $512\times512$  patches and generate the final flare-free images. All experiments were performed on two NVIDIA RTX 3090 GPUs with 24GB memory.
\paragraph{\bf GoodGame}
This team proposes an efficient flare removal network, based on Restormer\cite{Zamir2021Restormer}. In the model, they use Flare-Aware Transformer Blocks to capture the Flare in the image, and the composition structure is similar to that of Restormer. Residual connections are also used in the model so that the model only needs to learn the changes in the flare and does not need to reconstruct the entire image. The model is also efficient enough to infer large-resolution images. Fig.~\ref{fig:GoodGame} shows the framework of the entire model.

They trained 300,000 iterations to take the model to convergence. Progressive learning was used during the training process, from the initial batch size of 16 and patch size (resize) of 128, to the final batch size of 2 and patch size of 384. During the training process, gradually increase the patch size of the image and reduce the batch size, so that the model can learn more details of the image. They choose \texttt{AdamW} as an optimizer, set the initial learning rate to $3 \times 10^{-4}$, and introduce a weight decay of $1e-4$. At the same time, they adopted the cosine annealing learning rate scheduler (\texttt{CosineAnnealingLR}), where $T_{\max}$ is set to $500$ and the minimum learning rate is set to $1e-6$.In terms of loss, they used L1 loss, Fourier L1 loss, and Lpips loss, with Lpips loss accounting for the largest proportion.

\begin{figure}[!htp]
    \centering
    \includegraphics[width=\linewidth]{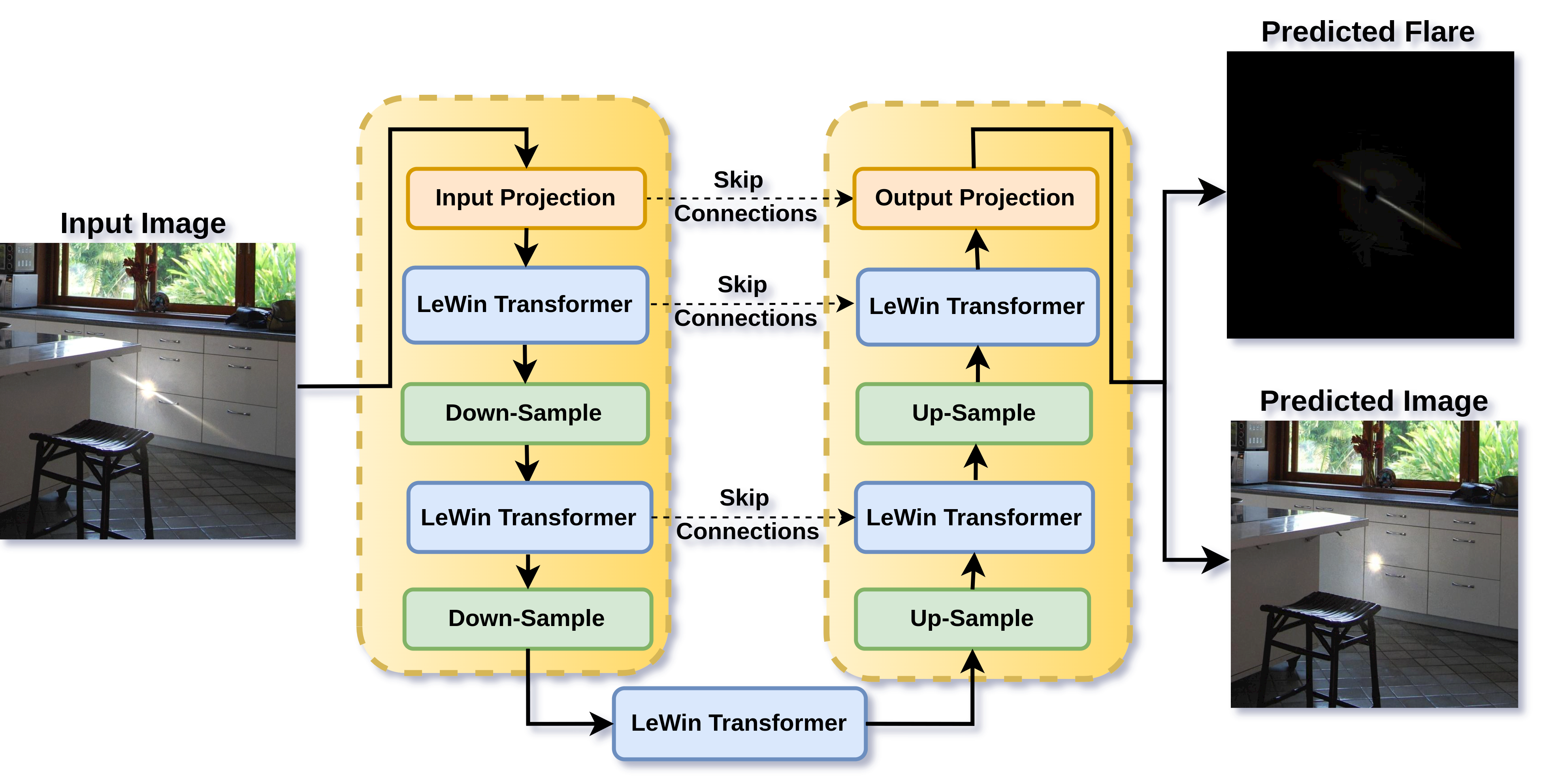}
    \caption{The network architecture of FADU-Net $/$ IIT-RPR team.}
    \label{Fadu-net}
    \vspace{-5mm}
\end{figure}

\paragraph{\bf IIT-RPR}
This team designs a method, based on U-former model architecture. FADU-Net shown
in Figure~\ref{Fadu-net}, trained from the ground up. The synthesis of training images involves utilizing Flickr24K as background images and incorporating 5k scattering flare images from Flare7K. An innovative night data augmentation strategy (Night Data Aug) is implemented for background images, featuring four modes,
randomly selected for each image during training. The provided 
competition's images are also used as the validation dataset during the training of the architecture. The loss function is a combination of L1 loss and perceptual
loss, with distinct weights assigned to areas inside and outside the flare. \\
During training, a patch size of 512 is utilized, and the Adam optimizer is employed with an initial
learning rate of 0.0001. This team conducts training for 1200K iterations and observes that extending the training
duration may further enhance results. Although, the proposed pipeline demonstrates its efficacy by achieving
a PSNR of 20.66 and LPIPS of 0.1926 on the MIPI challenge’s test dataset, showcasing the team’s adeptness
in addressing the complex task of nighttime flare removal. It outputs the two images i.e., the predicted flare and the predicted image.
The predicted flare image shows the reflective and scattering flare present in the input image. And predicted image is the output image without the flares containing only the light source. This architecture is trained on the 11GB NVIDIA GeForce RTX 2080 Ti GPU for 6 days and 13 hours.

\paragraph{\bf Hp\_zhangGeek}
This team designs a conditional variational autoencoder (CVAE) \cite{sohn2015learning} for removing nighttime flares. Specifically, for the nighttime flare removal task, CVAEs can contribute significantly due to their ability to model complex data distributions and generate high-quality, diverse outputs conditioned on given inputs. They also designed an Adaptive Normalization Module (ANM) to enhance the details of input features.

\begin{figure}[!h]
    \centering
    \includegraphics[width=1\linewidth]{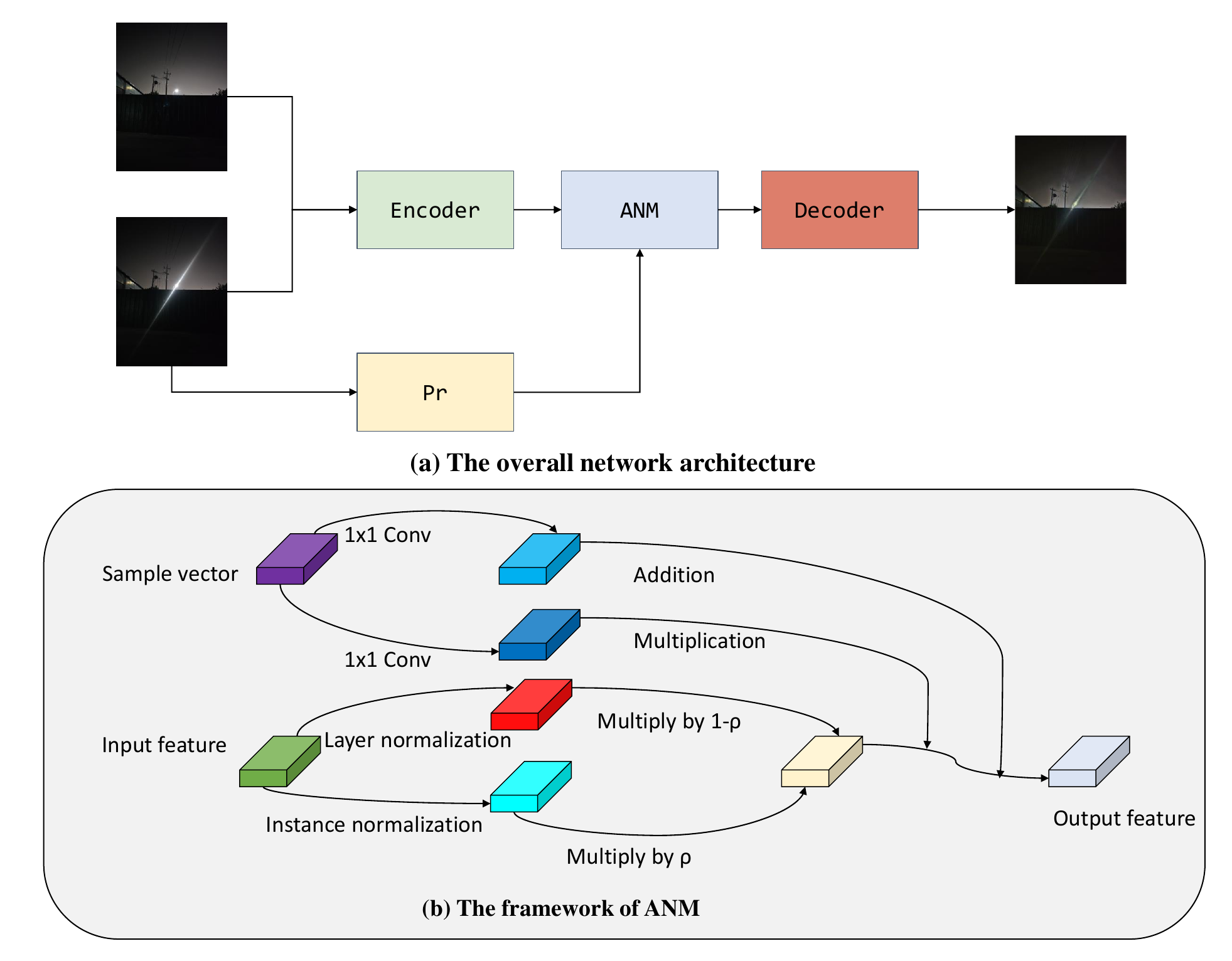}
    \vspace{-0.5cm}
    \caption{The overall network architecture of HP\_zhang Geek team.}
    \label{fig-zhp}
    % \vspace{-0.5cm}

\end{figure}

As depicted in Fig. \ref{fig-zhp} (a), they adopt U-Net architecture \cite{ronneberger2015u} in the encoder that progressively downsamples the image into a more compact representation. The Prior network (abbreviated as Pr in Fig. \ref{fig-zhp} (a)) shares the same structure as the encoder. This network is designed to learn a prior distribution of the latent variables. In the network, the prior network models the distribution of latent variables that is from the flare-corrupted images. Inspired by PUIE-Net \cite{fu2022uncertainty} and U-GAT-IT \cite{kim2019u}, the adaptive normalization module (ANM) takes the latent representation from the encoder and refines it. As shown in Fig. \ref{fig-zhp} (b), the adaptive normalization module involves adjusting the feature distribution of the latent representation to a state that is more conducive to generating a clean image without flares. Using the refined latent representation from the ANM, the decoder network reconstructs the image. The decoder effectively reverses the encoding process, upscaling the latent representation back to the original image dimensions to reproduce the image without the lens flare. 
 
This team only used the 600 image pairs for training. In cases where the dataset is relatively small, they adopt a $90/10$ split to maximize the amount of training data while still having a validation set to monitor overfitting and performance. They resize the images to $256 \times 256$ and apply horizontal flipping, vertical flipping, and rotation randomly for data augmentation. The batch size is $16$. They train the network for $700$ epochs. For the training phase, the loss function for training is formulated as follows:
\begin{equation} \label{eq:1}
\mathcal{L}_{\mathrm{total}}=\mathcal{L}_{\mathrm{re}}+\alpha\mathcal{L}_{\mathrm{kl}}+\beta\mathcal{L}_{\mathrm{per}},   
\end{equation}
where $\mathcal{L}_{\mathrm{re}}$ is the reconstruction loss, $\mathcal{L}_{\mathrm{kl}}$ is the KL divergence, and $\mathcal{L}_{\mathrm{per}}$ is the perceptual loss, $\alpha$ and  $\beta$ are learnable parameters.

The experimental setup for the computational framework was implemented on an Ubuntu-based workstation equipped with a single NVIDIA RTX 4090 card. The duration of the model training, inclusive of the validation phase, spanned approximately 16 hours. The inference time of the model is approximately 6.3 milliseconds per image. It is pertinent to note that the inference protocol entailed sampling the model 20 times for each image. The final output was derived by computing the mean across these 20 samples, ensuring robustness and stability in the generated results. This methodical approach to inference underscores the model's efficacy in handling the variability inherent in the data, thereby contributing to the reliability of the outcomes.

\paragraph{\bf Lehaan}
This team utilizes the Uformer model \cite{wang2022uformer} as a flare-erasing module coupled with AOT-GAN for image inpainting. Uformer employs a hierarchical encoder-decoder structure akin to UNet but substitutes convolutional layers with Transformer blocks. Key aspects of Uformer include the Locally-enhanced window (LeWin) Transformer block for localized context capture and the Multi-scale restoration modulator for feature adjustment at various scales. AOT-GAN (Aggregation of Contextual Transformation - GAN) enhances context reasoning through AOT blocks in the generator, facilitating the aggregation of contextual transformations from different image areas for accurate inpainting. AOT blocks are a novel approach for convolutional neural networks designed to enhance context reasoning. They achieve this by splitting a large kernel into smaller ones, each specializing in a specific number of output channels. These sub-kernels then analyze the input using different dilation rates, allowing them to focus on varying areas of the image. Finally, the outputs from all sub-kernels are merged, enabling the AOT block to consider the input from various perspectives and capture richer contextual information. This approach has shown promise in improving tasks like image inpainting.

While the Uformer model can sufficiently remove the flare, it also inadvertently removes the pixels that were behind the flares. Hence, an inpainting module is used to inpaint back the image that was removed too.
To specify the region required for inpainting, image differencing followed by thresholding is done in order to get the regions affected by UFormer. This is given as the mask to AOT-GAN while inpainting.

The complexity of the Uformer method comprises several stages:
Self Attention: Time complexity - $O(n^2d^2)$, Space complexity - $O(n^2d)$.
Feed-Forward Network: Time complexity - $O(2nd^2)$, Space complexity - $O(nd)$.
Layer Normalization and Residual Connection: Constant time and space complexity - $O(1)$. Multi-Head Attention (MHA): Time complexity - $O(nh^2d^2)$. Total time complexity:$O(n^2hd^2)$, Space Complexity $O(n^2d + nhd)$.

For model training
Mini-Batch size: 8, Epochs: 1000, Training workers: 4, Evaluation workers: 4, Dataset: Flare7k++, Optimizer: AdamW, learning rate = 10e-3 and default decay parameters, Weight decay: 0.02, GPU: NVIDIA RTX 4050 Laptop GPU.
\section{Conclusions}
In this report, we review and summarize the methods and results of MIPI 2024 challenge on Nighttime Flare Removal.
The participants have made significant contributions to this challenging track, and we express our gratitude for the dedication of each participant.
{
    \small
    \bibliographystyle{ieeenat_fullname}
    \bibliography{main}

\begin{thebibliography}{29}
\providecommand{\natexlab}[1]{#1}
\providecommand{\url}[1]{\texttt{#1}}
\expandafter\ifx\csname urlstyle\endcsname\relax
  \providecommand{\doi}[1]{doi: #1}\else
  \providecommand{\doi}{doi: \begingroup \urlstyle{rm}\Url}\fi

\bibitem[Chen et~al.(2022)Chen, Chu, Zhang, and Sun]{chen2022simple}
Liangyu Chen, Xiaojie Chu, Xiangyu Zhang, and Jian Sun.
\newblock Simple baselines for image restoration.
\newblock In \emph{European Conference on Computer Vision}, 2022.

\bibitem[Dai et~al.(2022)Dai, Li, Zhou, Feng, and Loy]{dai2022flare7k}
Yuekun Dai, Chongyi Li, Shangchen Zhou, Ruicheng Feng, and Chen~Change Loy.
\newblock Flare7k: A phenomenological nighttime flare removal dataset.
\newblock In \emph{Thirty-sixth Conference on Neural Information Processing Systems Datasets and Benchmarks Track}, 2022.

\bibitem[Dai et~al.(2023{\natexlab{a}})Dai, Li, Zhou, Feng, Luo, and Loy]{dai2023flare7k++}
Yuekun Dai, Chongyi Li, Shangchen Zhou, Ruicheng Feng, Yihang Luo, and Chen~Change Loy.
\newblock Flare7k++: Mixing synthetic and real datasets for nighttime flare removal and beyond.
\newblock \emph{arXiv preprint arXiv:2306.04236}, 2023{\natexlab{a}}.

\bibitem[Dai et~al.(2023{\natexlab{b}})Dai, Li, Zhou, Feng, Zhu, Sun, Sun, Loy, Gu, Liu, et~al.]{dai2023mipi}
Yuekun Dai, Chongyi Li, Shangchen Zhou, Ruicheng Feng, Qingpeng Zhu, Qianhui Sun, Wenxiu Sun, Chen~Change Loy, Jinwei Gu, Shuai Liu, et~al.
\newblock Mipi 2023 challenge on nighttime flare removal: Methods and results.
\newblock In \emph{IEEE Conference on Computer Vision and Pattern Recognition}, 2023{\natexlab{b}}.

\bibitem[Dai et~al.(2023{\natexlab{c}})Dai, Luo, Zhou, Li, and Loy]{dai2023reflective}
Yuekun Dai, Yihang Luo, Shangchen Zhou, Chongyi Li, and Chen~Change Loy.
\newblock Nighttime smartphone reflective flare removal using optical center symmetry prior.
\newblock In \emph{IEEE Conference on Computer Vision and Pattern Recognition}, 2023{\natexlab{c}}.

\bibitem[Fu et~al.(2022)Fu, Wang, Huang, Ding, and Ma]{fu2022uncertainty}
Zhenqi Fu, Wu Wang, Yue Huang, Xinghao Ding, and Kai-Kuang Ma.
\newblock Uncertainty inspired underwater image enhancement.
\newblock In \emph{European Conference on Computer Vision}. Springer, 2022.

\bibitem[Kim et~al.(2019)Kim, Kim, Kang, and Lee]{kim2019u}
Junho Kim, Minjae Kim, Hyeonwoo Kang, and Kwang~Hee Lee.
\newblock U-gat-it: Unsupervised generative attentional networks with adaptive layer-instance normalization for image-to-image translation.
\newblock In \emph{International Conference on Learning Representations}, 2019.

\bibitem[Kingma and Ba(2014)]{kingma2014adam}
Diederik~P Kingma and Jimmy Ba.
\newblock Adam: A method for stochastic optimization.
\newblock \emph{arXiv preprint arXiv:1412.6980}, 2014.

\bibitem[Lan and Chen(2023)]{lan2023tackling}
Fengbo Lan and Chang~Wen Chen.
\newblock Tackling scattering and reflective flare in mobile camera systems: A raw image dataset for enhanced flare removal.
\newblock \emph{arXiv preprint arXiv:2307.14180}, 2023.

\bibitem[Luo(2023)]{luo2023image}
Ziwei Luo.
\newblock Image restoration with mean-reverting stochastic differential equations.
\newblock In \emph{International Conference on Machine Learning}, 2023.

\bibitem[Mao et~al.(2021)Mao, Liu, Shen, Li, and Wang]{mao2021deep}
Xintian Mao, Yiming Liu, Wei Shen, Qingli Li, and Yan Wang.
\newblock Deep residual fourier transformation for single image deblurring.
\newblock \emph{arXiv preprint arXiv:2111.11745}, 2021.

\bibitem[Ronneberger et~al.(2015)Ronneberger, Fischer, and Brox]{ronneberger2015u}
Olaf Ronneberger, Philipp Fischer, and Thomas Brox.
\newblock U-net: Convolutional networks for biomedical image segmentation.
\newblock In \emph{Medical Image Computing and Computer-Assisted Intervention}. Springer, 2015.

\bibitem[Simonyan and Zisserman(2015)]{simonyan2014very}
K Simonyan and A Zisserman.
\newblock Very deep convolutional networks for large-scale image recognition.
\newblock In \emph{3rd International Conference on Learning Representations (ICLR 2015)}. Computational and Biological Learning Society, 2015.

\bibitem[Sohn et~al.(2015)Sohn, Lee, and Yan]{sohn2015learning}
Kihyuk Sohn, Honglak Lee, and Xinchen Yan.
\newblock Learning structured output representation using deep conditional generative models.
\newblock \emph{Advances in Neural Information Processing Systems}, 2015.

\bibitem[Sun et~al.(2023{\natexlab{a}})Sun, Yang, Li, Zhou, Feng, Dai, Sun, Zhu, Loy, Gu, et~al.]{sun2023mipi}
Qianhui Sun, Qingyu Yang, Chongyi Li, Shangchen Zhou, Ruicheng Feng, Yuekun Dai, Wenxiu Sun, Qingpeng Zhu, Chen~Change Loy, Jinwei Gu, et~al.
\newblock Mipi 2023 challenge on rgbw remosaic: Methods and results.
\newblock In \emph{IEEE Conference on Computer Vision and Pattern Recognition}, 2023{\natexlab{a}}.

\bibitem[Sun et~al.(2023{\natexlab{b}})Sun, Yang, Li, Zhou, Feng, Dai, Sun, Zhu, Loy, Gu, et~al.]{sun2023mipi2}
Qianhui Sun, Qingyu Yang, Chongyi Li, Shangchen Zhou, Ruicheng Feng, Yuekun Dai, Wenxiu Sun, Qingpeng Zhu, Chen~Change Loy, Jinwei Gu, et~al.
\newblock Mipi 2023 challenge on rgbw fusion: Methods and results.
\newblock In \emph{IEEE Conference on Computer Vision and Pattern Recognition}, pages 2870--2876, 2023{\natexlab{b}}.

\bibitem[Vasluianu and Timofte(2022)]{9897883}
Florin Vasluianu and Radu Timofte.
\newblock Efficient video enhancement transformer.
\newblock In \emph{IEEE International Conference on Image Processing}, 2022.

\bibitem[Wang et~al.(2018)Wang, Liu, Zhu, Tao, Kautz, and Catanzaro]{wang2018pix2pixHD}
Ting-Chun Wang, Ming-Yu Liu, Jun-Yan Zhu, Andrew Tao, Jan Kautz, and Bryan Catanzaro.
\newblock High-resolution image synthesis and semantic manipulation with conditional gans.
\newblock In \emph{IEEE Conference on Computer Vision and Pattern Recognition}, 2018.

\bibitem[Wang et~al.(2004)Wang, Bovik, Sheikh, and Simoncelli]{ssim}
Zhou Wang, Alan~C Bovik, Hamid~R Sheikh, and Eero~P Simoncelli.
\newblock Image quality assessment: from error visibility to structural similarity.
\newblock \emph{IEEE Transactions on Image Processing}, 13\penalty0 (4):\penalty0 600--612, 2004.

\bibitem[Wang et~al.(2022{\natexlab{a}})Wang, Cun, Bao, Zhou, Liu, and Li]{Uformer}
Zhendong Wang, Xiaodong Cun, Jianmin Bao, Wengang Zhou, Jianzhuang Liu, and Houqiang Li.
\newblock Uformer: A general u-shaped transformer for image restorationn.
\newblock In \emph{IEEE Conference on Computer Vision and Pattern Recognition}, 2022{\natexlab{a}}.

\bibitem[Wang et~al.(2022{\natexlab{b}})Wang, Cun, Bao, Zhou, Liu, and Li]{wang2022uformer}
Zhendong Wang, Xiaodong Cun, Jianmin Bao, Wengang Zhou, Jianzhuang Liu, and Houqiang Li.
\newblock Uformer: A general u-shaped transformer for image restoration.
\newblock In \emph{IEEE Conference on Computer Vision and Pattern Recognition}, 2022{\natexlab{b}}.

\bibitem[Wu et~al.(2021{\natexlab{a}})Wu, He, Xue, Garg, Chen, Veeraraghavan, and Barron]{how_to}
Yicheng Wu, Qiurui He, Tianfan Xue, Rahul Garg, Jiawen Chen, Ashok Veeraraghavan, and Jonathan~T. Barron.
\newblock How to train neural networks for flare removal.
\newblock In \emph{IEEE International Conference on Computer Vision}, 2021{\natexlab{a}}.

\bibitem[Wu et~al.(2021{\natexlab{b}})Wu, He, Xue, Garg, Chen, Veeraraghavan, and Barron]{wu2021train}
Yicheng Wu, Qiurui He, Tianfan Xue, Rahul Garg, Jiawen Chen, Ashok Veeraraghavan, and Jonathan~T Barron.
\newblock How to train neural networks for flare removal.
\newblock In \emph{IEEE International Conference on Computer Vision}, 2021{\natexlab{b}}.

\bibitem[Zamir et~al.(2022)Zamir, Arora, Khan, Hayat, Khan, and Yang]{Zamir2021Restormer}
Syed~Waqas Zamir, Aditya Arora, Salman Khan, Munawar Hayat, Fahad~Shahbaz Khan, and Ming-Hsuan Yang.
\newblock Restormer: Efficient transformer for high-resolution image restoration.
\newblock In \emph{IEEE Conference on Computer Vision and Pattern Recognition}, 2022.

\bibitem[Zeng et~al.(2022)Zeng, Fu, Chao, and Guo]{zeng2022aggregated}
Yanhong Zeng, Jianlong Fu, Hongyang Chao, and Baining Guo.
\newblock Aggregated contextual transformations for high-resolution image inpainting.
\newblock \emph{IEEE Transactions on Visualization and Computer Graphics}, 2022.

\bibitem[Zhang et~al.(2018)Zhang, Isola, Efros, Shechtman, and Wang]{lpips}
Richard Zhang, Phillip Isola, Alexei~A. Efros, Eli Shechtman, and Oliver Wang.
\newblock The unreasonable effectiveness of deep features as a perceptual metric.
\newblock In \emph{IEEE Conference on Computer Vision and Pattern Recognition}, 2018.

\bibitem[Zhao et~al.(2022)Zhao, Zhang, Hong, Xu, Yang, and Wang]{fclgan}
Suiyi Zhao, Zhao Zhang, Richang Hong, Mingliang Xu, Yi Yang, and Meng Wang.
\newblock Fcl-gan: A lightweight and real-time baseline for unsupervised blind image deblurring.
\newblock In \emph{30th ACM International Conference on Multimedia}, 2022.

\bibitem[Zhou et~al.(2023)Zhou, Liang, Chen, Huang, Yang, and Li]{zhou2023improving}
Yuyan Zhou, Dong Liang, Songcan Chen, Sheng-Jun Huang, Shuo Yang, and Chongyi Li.
\newblock Improving lens flare removal with general-purpose pipeline and multiple light sources recovery.
\newblock In \emph{IEEE International Conference on Computer Vision}, 2023.

\bibitem[Zhu et~al.(2023)Zhu, Sun, Dai, Li, Zhou, Feng, Sun, Loy, Gu, Yu, et~al.]{zhu2023mipi}
Qingpeng Zhu, Wenxiu Sun, Yuekun Dai, Chongyi Li, Shangchen Zhou, Ruicheng Feng, Qianhui Sun, Chen~Change Loy, Jinwei Gu, Yi Yu, et~al.
\newblock Mipi 2023 challenge on rgb+ tof depth completion: Methods and results.
\newblock In \emph{IEEE Conference on Computer Vision and Pattern Recognition}, 2023.

\end{thebibliography}
}

\appendix

\section{Teams and Affiliations}
\label{append:teams}

\subsection*{\bf MiAlgo\_AI}
{\bf Title:} PPDN - Progressive Perception Diffusion Network for Nighttime Flare Removal\\
{\bf Members:}\\
Lize Zhang$^1$ (\href{zhanglize@xiaomi.com}{zhanglize@xiaomi.com})\\
Shuai Liu$^1$\quad Chaoyu Feng$^1$\quad Luyang Wang$^1$\quad Shuan Chen$^1$\quad Guangqi Shao$^1$\quad Xiaotao Wang$^1$\quad Lei Lei$^1$\\
{\bf Affiliations:}\\
$^1$ Xiaomi Inc., China

\subsection*{\bf BigGuy}
{\bf Title:} Progressive Inference and Mask Loss for Nighttime Flare Removal\\
{\bf Members:}\\
Qirui Yang$^1$ (\href{yangqirui@tju.edu.cn}{yangqirui@tju.edu.cn})\\
Qihua Cheng$^2$  \quad Zhiqiang Xu$^2$ \quad Yihao Liu$^3$ \quad  Huanjing Yue$^1$ \quad Jingyu Yang$^1$ \\
{\bf Affiliations:}\\
$^1$ School of Electrical and Information Engineering, Tianjin University, China \\
$^2$ Shenzhen MicroBT Electronics Technology Co., Ltd,  China \\
$^3$ Shanghai Artificial Intelligence Laboratory, China.

\subsection*{\bf SFNet-FR}
{\bf Title:} SFNet - A Spatial-Frequency Domain Neural Network for Image Lens Flare Removal\\
{\bf Members:}\\
Florin-Alexandru Vasluianu$^1$ \\(\href{florin-alexandru.vasluianu@uni-wuerzburg.de}{florin-alexandru.vasluianu@uni-wuerzburg.de})\\
Zongwei Wu$^1$\quad George Ciubotariu$^1$\quad Radu Timofte$^1$\\
{\bf Affiliations:}\\
$^1$ Computer Vision Lab, CAIDAS \& IFI, University of W\"urzburg, Germany

\subsection*{\bf LVGroup\_HFUT}
{\bf Title:} Exploring nighttime flare removal with different distributions\\
{\bf Members:}\\
Zhao Zhang$^1$ (\href{cszzhang@gmail.com}{cszzhang@gmail.com})\\
Suiyi Zhao$^1$\quad Bo wang$^1$\quad Zhichao Zuo$^1$\quad Yanyan Wei\\
{\bf Affiliations:}\\
$^1$ Laboratory for Multimedia Computing, Hefei University of Technology, China

\subsection*{\bf CILAB-IITMadras}
{\bf Title:} Efficient flare removal using ensembling methods \\
{\bf Members:}\\
Kuppa Sai Sri Teja$^1$ (\href{saisriteja@detecttechnologies.com}{saisriteja@detecttechnologies.com})\\
Jayakar Reddy$^2$ 
Girish Rongali$^2$
Kaushik Mitra$^2$ \\
{\bf Affiliations:}\\
$^1$ Detect Technologies Pvt Ltd, India \\
$^2$ Indian Institute of Technology, Madras, India. 

\subsection*{\bf Xdh-Flare}
{\bf Title:} Augmenting Dataset across Domains for Nighttime Flare Removal \\
{\bf Members:}\\
Zhihao Ma$^1$ (\href{23021211614@stu.xidian.edu.cn}{23021211614@stu.xidian.edu.cn})\\
Yongxu Liu$^1$ \\
{\bf Affiliations:}\\
$^1$ Xidian University, China

\subsection*{\bf Fromhit}
{\bf Title:} Nighttime Flare Removal Network based on NAFNet\\
{\bf Members:}\\
Wanying Zhang$^1$ (\href{22b903084@stu.hit.edu.cn}{22b903084@stu.hit.edu.cn})\\
Wei Shang$^1$ \\
{\bf Affiliations:}\\
$^1$ Harbin Institute of Technology, China

\subsection*{\bf UformerPlus}
{\bf Title:} Enhancing Nighttime Flare Removal with Uformer and Frequency-based Residual Blocks\\
{\bf Members:}\\
Yuhong He$^1$ (\href{madeline0728@163.com}
{madeline0728@163.com})\\
Long Peng$^2$ \hspace{4pt} Zhongxin Yu$^3$ \hspace{4pt} Shaofei Luo$^3$ \\
{\bf Affiliations:}\\
$^1$ Northeastern University, China \\
$^2$ University of Science and Technology of China, China \\
$^3$ Fujian Normal University, China

\subsection*{\bf GoodGame}
{\bf Title:} Efficient Flare Removal Network based on Restormer\\
{\bf Members:}\\
Jian Wang$^1$ (\href{jwang4@snapchat.com}{jwang4@snapchat.com})\\
Yuqi Miao$^2$ \hspace{4pt}Baiang Li$^3$ \hspace{4pt} Gang Wei$^2$ \\
{\bf Affiliations:}\\
$^1$ Snap Inc., USA \\
$^2$ Tongji University, China \\ 
$^3$ Hefei University of Technology, China

\subsection*{\bf IIT-RPR}
{\bf Title:} Flare Annihilation and Deflare Unit (FADU) Network\\
{\bf Members:}\\
Rakshank Verma$^1$ (\href{rakshankverma130250@gmail.com}{rakshankverma130250@gmail.com})\\
Ritik Maheshwari$^1$ \hspace{4pt}Rahul Tekchandani$^1$ \hspace{4pt} Praful Hambarde$^2$ \hspace{4pt} Satya Narayan Tazi$^1$ \hspace{4pt} Santosh Kumar Vipparthi$^2$ \hspace{4pt} Subrahmanyam Murala$^3$\\
{\bf Affiliations:}\\
$^1$ GEC Ajmer, India \\ 
$^2$ CVPR Lab IIT Ropar, India \\ 
$^3$ SCSS Trinity College Dublin, Ireland

\subsection*{\bf Hp\_zhangGeek}
{\bf Title:} Enhancing Nighttime Flare Removal with CVAE and Adaptive Normalization Module\\
{\bf Members:}\\
Haopeng Zhang$^1$ (\href{zhanghaopeng0606@gmail.com}
{zhanghaopeng0606@gmail.com})\\
Yingli Hou$^2$ \hspace{4pt} Mingde Yao$^3$ \\
{\bf Affiliations:}\\
$^1$ Faculty of Robot Science and Engineering, Northeastern University, China \\ 
$^2$ Software College, Northeastern University, China \\ 
$^3$ The Chinese University of Hong Kong, China

\subsection*{\bf Lehaan}
{\bf Title:} Enhancing Nighttime Flare Removal with Uformer and AOT-GAN\\
{\bf Members:}\\
Levin M S$^1$ (\href{mslevin.active@gmail.com}{mslevin.active@gmail.com})\\
Aniruth Sundararajan$^1$ \hspace{4pt} Hari Kumar A$^1$ \\
{\bf Affiliations:}\\
$^1$ Shiv Nadar University Chennai, India

% WARNING: do not forget to delete the supplementary pages from your submission 
% \input{sec/X_suppl}

\end{document}